# Deep learning-based Visual Measurement Extraction within an Adaptive Digital Twin Framework from Limited Data Using Transfer Learning

[1]Mehrdad Shafiei Dizaji, Ph.D., Postdoctoral Research Associate,

Department of Engineering Systems and Environment, University of Virginia, 151 Engineer's Way, Charlottesville, VA, 22904, USA, Phone:(434)987-9780, Email: ms4qg@virginia.edu

Devin K. Harris, Ph.D., Professor,

Department of Engineering Systems and Environment, University of Virginia, 151 Engineer's Way, Charlottesville, VA, 22904, USA, Phone: (434) 924-6373; Email: dharris@virginia.edu

Zahra Zhiyanpour, Research Assistance,

Department of Engineering Systems and Environment, University of Virginia, 151 Engineer's Way, Charlottesville, VA, 22904, USA, Phone:(434)987-9780, Email: pwd6qv@virginia.edu

Aya Yehia, Research Assistance,

Department of Engineering Systems and Environment, University of Virginia, 151 Engineer's Way, Charlottesville, VA, 22904, USA, Phone:(434)987-9780, Email: ay8tb@virginia.edu

**Abstract**

Digital Twins technology is revolutionizing decision-making in scientific research by integrating models and simulations with real-time data. Unlike traditional Structural Health Monitoring methods, which rely on computationally intensive Digital Image Correlation and have limitations in real-time data integration, this research proposes a novel approach using Artificial Intelligence. Specifically, Convolutional Neural

---

[1] Corresponding author, Phone:(434)987-9780, Email: ms4qg@virginia.edu

Networks are employed to analyze structural behaviors in real-time by correlating Digital Image Correlation speckle pattern images with deformation fields. Initially focusing on two-dimensional speckle patterns, the research extends to three-dimensional applications using stereo-paired images for comprehensive deformation analysis. This method overcomes computational challenges by utilizing a mix of synthetically generated and authentic speckle pattern images for training the Convolutional Neural Networks. The models are designed to be robust and versatile, offering a promising alternative to traditional measurement techniques and paving the way for advanced applications in three-dimensional modeling. This advancement signifies a shift towards more efficient and dynamic structural health monitoring by leveraging the power of Artificial Intelligence for real-time simulation and analysis.

# 1. Introduction

*1.1. Research Gap, Motivation and Problem Addressing*

In recent scholarly discourse, the notion of a DT represents a progressive extension of the SHM paradigm. This advancement encompasses a comprehensive approach, encompassing the entirety of a structure's life cycle. The DT paradigm elevates numerical models to serve as a virtual counterpart of the physical system. These models are synergistically enhanced by integrating data from diverse sources, facilitating insights into both present and prospective performance metrics of the structural system [1-4]. In this study, we explore the integration of a model-centric approach within the DT framework, specifically focusing on extracting visual measurements from image-based sensors using a learning algorithm based on a non-physics model. The primary inquiry of this research is to determine the feasibility of using AI algorithms, trained with visual sensing data, to identify and quantify full-field measurements. This method's potential is particularly significant in the context of the DT framework. It offers the prospect of creating a real-time digital representation of existing infrastructure systems, especially in scenarios where the initial condition of the structure is unknown. Such a digital representation is invaluable for initially assessing the structure's state, thereby enabling the measurement of the model's capacity based on the proposed non-physics-based model.

Real-time measurement represents the cornerstone of the DT framework [5-7]. The DT paradigm introduces a novel enhancement to the SHM framework, emphasizing the role of prototypes and simulations as central elements in decision-making processes. The design of these DTs necessitates a high degree of agility to facilitate the swift and seamless incorporation of empirical results into a dynamic feedback loop with the modeled representations of the structure [8]. In SHM systems, techniques like DIC are commonly employed to analyze the comprehensive behavior of structures. Nevertheless, these methods typically incur substantial computational expenses, which limit the bidirectional exchange between empirical data and simulations. By adopting real-time analysis of structures, the need to solve complex system equations is bypassed. This can be achieved through the implementation of non-physics-based representations of structures, utilizing advanced AI methodologies, including deep learning techniques. Therefore, a primary objective of this proposed approach is to leverage recent advances in the field of AI, namely deep learning to develop a real time full-field deformation measurement models that can inspect and extract deformation fields of a large area of an entire structure in a matter of seconds, which would otherwise take hours of manual work. Such a model can measure and extract full-field deformation fields rapidly to determine whether the condition performance has changed since the last evaluation. The exploration of the proposed technology innovations enables the following research questions to be answered: 1) To what extent can AI-generated models accurately simulate the intricate condition assessment that are typically obtained through experimentation on structural systems? 2) To what extent does the development of these AI formulated models facilitate the real-time analysis enabling bi-directional interaction between simulation and experimentation, which is crucial for the creation of a large-scale structural DT?

*1.2. Image-based Structural Health Monitoring*

In recent years, DIC techniques have become increasingly popular in the field of physical infrastructure education, as they offer a non-contact measurement method for complex environments. The ability of DIC to provide operational response measurements without disrupting services presents a unique opportunity in the field [9]. The DIC method is a widely used photogrammetric technique that involves the use of image correlation algorithms to detect and monitor the movement of patterns in a series of images of a specimen undergoing deformation [10]. Unlike discrete sensor measurements, DIC offers full-field 3D surface deformation measurements that are consistent with the spatial distribution of results obtained through finite element analysis [11-14]. DIC facilitates the tracking and correlation of motion patterns on the surface of a specimen experiencing deformation. This technique proficiently ascertains comprehensive surface displacements and strains, yielding results that are comparable to those derived from a finite element model [15-20]. The methodology involves segmenting the images into intricate grids or subsets, serving as reference points. This is followed by a template matching and tracking procedure, wherein the movement of points on the specimen's surface is identified by optimizing the pattern correlation of pixels across successive images. Monitoring these movements over time furnishes comprehensive surface deformation data. Subsequently, this information can be utilized to deduce other structural responses, like strains, via post-processing techniques. The process of tracking the pixel-by-pixel movement within subsets from image to image is similar to the finite element method and enables the determination of full-field deformation across the specimen surface. However, the computational demands of DIC make its application in real-time environments challenging. In recent studies, the integration of DIC into a topology optimization framework has been proposed to detect hidden damage at the element level within simple structural members [11-13]. The authors have demonstrated the potential of DIC for characterizing component-level damage, but the long solution times remain a hindrance to the development of a true DT with real-time feedback capabilities. Moreover, DIC requires the selection of various parameters to estimate displacement accurately. One crucial parameter is the subset size, which should be optimized to balance accuracy and computational efficiency. In addition, other parameters such as the interpolation method, image filtering, and correlation function also significantly affect the accuracy of deformation measurements. The correlation function must be selected to ensure it provides the best correlation between the reference and deformed images. Furthermore , image filtering is necessary to remove the noise from the images, while the interpolation method is essential to determine the displacement of the specimen between the grid points. Accurate parameter selection is essential for achieving reliable displacement measurements, and several techniques such as optimization algorithms and sensitivity analysis can be used to optimize these parameters. To provide an alternative approach, a deep learning-based DIC method is proposed, which is a learning-based deformation measurement method. Hence, the aim of this proposed approach is to harness the advancements in artificial intelligence, particularly deep learning, to create real-time full-field deformation measurement models. These models allow for rapid inspection and extraction of deformation fields across an entire structure in just minutes, compared to the hours of manual work currently required. Plus, the method offers a significant advantage over traditional DIC methods, as it enables simultaneous estimation of displacement by learning from multiple steps. This learning-based deformation measurement method enables end-to-end learning by taking input data, such as images, and desired goals, such as the target displacement values, to produce accurate results. Unlike the DIC method, the deep learning based DIC method does not require the identification of optimal parameters, such as subset size, since it estimates non-linear displacement using the input image and target value for various deformations. Consequently, the Deep learning based DIC method can accurately estimate displacement without the need for identifying optimal parameters as required in the DIC method. By using a learning-based approach, the deep learning based DIC method has the potential to improve the accuracy and efficiency of deformation measurements and expand the application of DIC in various fields.

*1.3. Deep CNN based DIC analyzer*

In deep learning, a CNNs is a class of artificial neural network (ANN), most commonly applied to analyze visual imagery [21]. An analysis of CNNs reveals certain parallels with DIC algorithms. Both methodologies employ kernel-based processes: DIC uses subset correlation calculations, whereas CNNs utilize convolution operations. The process of peak searching in DIC bears resemblance to the max-pooling layer employed in CNNs. Nonetheless, a notable distinction exists in the computation of the correlation criterion. DIC implements a complex nonlinear function, whereas CNNs derive feature maps through a linear computation of kernel values, subsequently augmented by an activation function. Through the use of multiple layers, CNN-based methods may provide improved performance compared to traditional DIC algorithms by capturing complex, nonlinear relationships between inputs and outputs. Min et al. (2019) [22] engineered a 3D CNN designed to extract features from both spatial and temporal domains in a series of image sets, aiming to predict the average displacement vector for each subset. The effectiveness of their model, however, was constrained by the limited size of the training dataset, which was expanded using data derived from experimental results. The anticipated prediction of the strain field was not accomplished, and the accuracy in forecasting the displacement field did not surpass that of conventional DIC algorithms. In their 2021 study, Boukhtache et al. [23] implemented deep learning techniques within the framework of DIC, taking cues from optical flow methodologies. They trained various CNNs, modeled after existing optical flow CNN architectures, using synthesized speckle image datasets to precisely predict sub-pixel level deformations or movements. The process involved initially employing a conventional correlation technique for identifying integer shifts, followed by the application of a CNN to detect sub-pixel deformations. Although this method yielded high accuracy, it remained a composite approach, incorporating subset division, post-filtering, and traditional correlation techniques. Despite previous efforts to integrate deep learning into DIC, there has yet to be a demonstration of reliable full-field strain field predictions solely based on deep learning methods. The study [23] presented a CNN deep learning-based methodology as a non-contact, vision-based solution for detecting deformation fields in SHM. The authors assessed the efficacy of this approach by training the model on images featuring simulated speckle patterns, and subsequently testing it on a dataset with varying characteristics, achieving high levels of accuracy.

Nonetheless, the proposed deep learning architecture exhibited constraints in fully leveraging spatial-temporal features from the full-field images. Moreover, the proposed method has not undergone fine-tuning or validation on an independent structure, which raises questions about its ability to generalize the trained model or estimate deformation fields data that were not part of the training dataset. These previous deep learning-based methods also showed little advantage over traditional DIC in terms of prediction accuracy, and did not have enough high-quality datasets to demonstrate their computational efficiency. Moreover, there is a noticeable lack of direct comparative analysis regarding the accuracy of displacement and strain fields between deep learning-based approaches and traditional DIC techniques. The practical effectiveness of deep learning enhanced DIC is yet to be distinctly understood, and prior studies have not explicitly articulated the rationale behind integrating deep learning into DIC methodologies. Given that DIC is an established method with readily available commercial software, the advantages of incorporating deep learning within this context remain ambiguous.

In an effort to circumvent the resource-intensive computational demands associated with employing DIC for deriving structural deformation fields from image sequences, a novel approach is suggested. This method, rooted in deep learning, seeks to reformulate the problem, offering a potentially more efficient solution. This solution is inspired using deep learning for estimating optical flow in the field of image processing and computer vision. Optical flow is the apparent relative motion of objects in consecutive frames of a video and its estimation has applications in object detection and tracking, robotic navigation, video stabilization and super-resolution, and 3D registration. A body of research exists on the use of deep learning for the estimation of optical flow from input sequences of natural images such as frames of a video with several benchmark datasets used by the computer vision community [23-29]. Research [30] indicates that methods based on CNNs have outperformed traditional optical flow techniques in accuracy and computational efficiency. The architecture of CNNs, characterized by multiple layers of convolution and deconvolution, complemented by appropriate pooling and activation functions [26], endows these networks with an exceptional capability. This allows for the precise recovery of optical flow fields at a sub-pixel level between pairs of images, effectively handling even large displacements [28]. However, the existing datasets to train the models contain sequences of natural scenes that are structurally different from the random speckle patterns used in the DIC technique for structural deformation sensing. The reformulation of the structural deformation fields using a similar approach requires the creation of training datasets containing speckle images as well as the development of deep learning architectures able to process 2D and 3D image sequences into corresponding deformation (and strain) fields.

The proposed deep learning-based model is inspired by the concept of DIC algorithms, which extract deformation fields from a sequence of images including a reference and deformed image. The deep learning model under discussion employs CNNs to trace pixel movements by harnessing the spatial-temporal characteristics of the images. This method enables the extraction of displacement and strain fields in a seamless, end-to-end fashion, negating the need for any interpolation or iterative processes. The objective of this approach is to deliver robust and precise predictions of comprehensive, high-resolution displacement and strain fields using sequences of speckle patterns. This model is designed for direct comparison with existing commercial DIC software. The anticipated advantages include enhanced strain prediction accuracy, comparable or superior accuracy in predicting small to moderate deformations, and a decrease in computation time. This reduction is particularly significant for applications requiring real-time measurements and predictions.

*1.4. Contributions*

This study focuses on overcoming the limitations of using DIC for real-time engagement in SHM. The DIC process is computationally intensive, making it difficult to provide real-time analysis. To address this issue, a deep learning-based approach is proposed to automatically identify deformation fields from image sequences. The suggested methodology employs a deep learning algorithm to facilitate an automated, non-destructive, image-centric approach to SHM. This algorithm is capable of delivering precise on-site evaluations of various engineering structures, thereby serving as an essential instrument for real-time surveillance. Utilizing two deep learning models, the approach bypasses traditional image processing techniques, directly converting image sequences into fundamental deformation fields. The deep learning models take images or videos of a structure as input and produce the associated deformation fields as output. The developed CNNs are tailored for 2D speckle patterns and their corresponding in-plane deformation fields, utilizing pairs of images. Future extensions of this model are planned to encompass 3D stereo images, necessitating modifications to handle stereo-paired speckle images, thereby addressing both in-plane and out-of-plane deformation fields. To overcome the challenge of the extensive data requirement for training deep learning models, the proposed method integrates artificially generated speckle pattern image pairs, featuring pre-set deformation patterns, with authentic speckle datasets obtained from DIC experiments. Distinguishing itself from previous efforts, this deep learning model directly forecasts the strain field from image pairs, eliminating the intermediate step of displacement prediction. This end-to-end methodology has shown multiple benefits over the conventional process of calculating the strain field through spatial derivatives of the displacement field. Conventional DIC typically utilizes spatial filtering in the computation of strain fields. However, this approach tends to diminish the spatial resolution of strain predictions and necessitates manual adjustment of filtering parameters. Notably, there is an absence of standardized guidelines for this parameter tuning process. The conceptualization of deep learning-based full-field visual measurement extraction is illustrated in Figure 1 schematically. The main contributions of this paper are:

1. The research introduces two innovative deep learning techniques for the extraction of deformation fields. These methods represent a significant advancement over current state-of-the-art approaches, offering real-time processing, autonomous operation, and enhanced accuracy in model generalization to new, limited datasets that have not been previously encountered.

2. The novel deep learning based DIC strategy introduced in this research negates the requirement for interpolation or iterative processes, which are standard in conventional DIC algorithms. This is achieved by directly deriving the strain field from the image input through an integrated, end-to-end methodology.

3. The presented method capitalizes on transfer learning to estimate deformation fields in a constrained, distinct dataset (referred to as *Dataset B*) [11], utilizing insights gleaned from a more extensive collection of simulated images (*Dataset A*) [23]. This application of transfer learning empowers the model to extend its learned knowledge, facilitating accurate predictions even with limited data availability.

4. Furthermore, it has demonstrated extrapolatory capability by being able to estimate deformation fields for unseen perspectives. The difference between generalizability and extrapolability is that the former focuses on interpolation within the training data while the latter focuses on making predictions for new, unseen data.

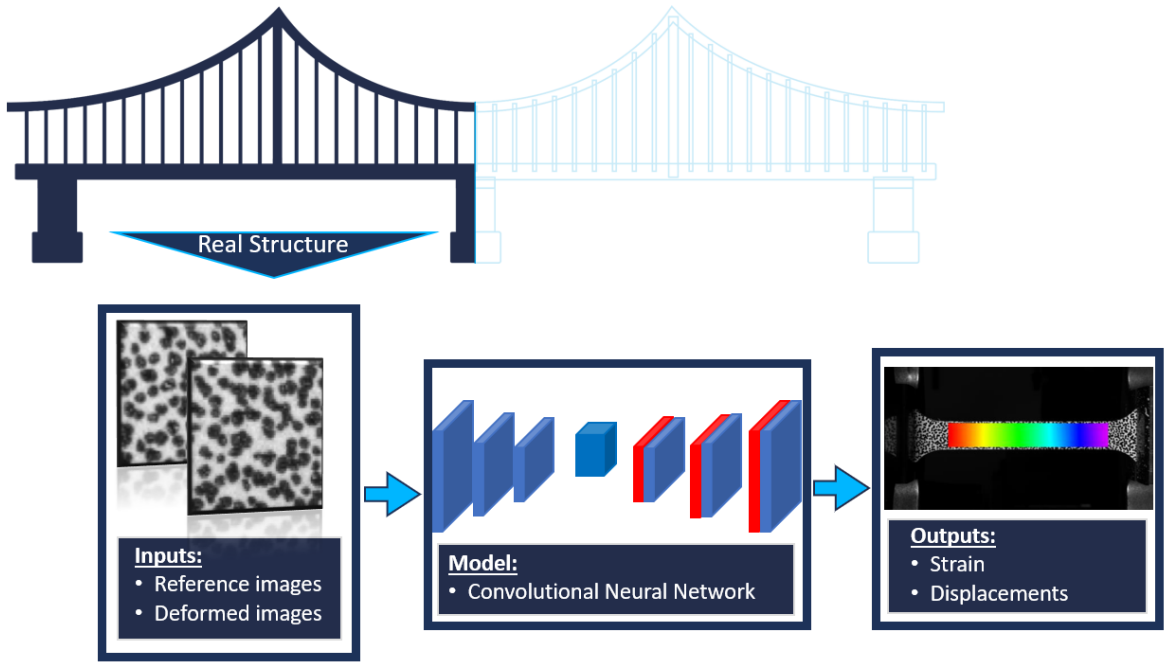

*Figure 1. Conceptualization of Deep learning-based full-field visual measurement extraction*

## 2. Deep Learning Architecture for full-field subpixel displacement extraction

This study concentrates on formulating deep CNNs based model to discern the correlation between sequences of DIC speckle pattern images and their associated deformation fields. The construction and execution of these CNNs are delineated, with the objective of autonomously ascertaining both displacement and strain fields from a series of image inputs. The overall workflow of the proposed CNN for the deformation fields extraction from 2D images is illustrated in Figure 2. The architecture of the network is displayed in Figure 3. In tackling the obstacle of acquiring substantial datasets required for training the advanced deep learning models, this research advocates for the use of a dual approach. This includes the employment of synthetically created speckle pattern image pairs that incorporate predetermined deformation patterns, coupled with authentic speckle datasets sourced from both newly conducted experiments and prior DIC experiments conducted by the authors [11].

## 2.1. Deep Learning Architecture trained with 2D DIC dataset

The current research introduces two CNNs – 'DisplacementCNN' and 'StrainCNN' – designed to deduce displacement and strain fields, respectively, from a pair of speckle images. For both networks, the inputs are speckle image pairs, each with a height '$h$' and width '$w$'. The dimensions of these input images may fluctuate due to the dynamic tracking of the Region of Interest (ROI). To ensure consistency in the input images during the inference stage, a pre-processing step is implemented. This step scales the images to the nearest multiples of 32 in both dimensions. Post-result acquisition, the images are re-scaled to their original dimensions for accurate displacement field analysis. Notably, the prediction of the strain field is not impacted by this rescaling process. It is noted that 'DisplacementCNN' and 'StrainCNN' exhibit marginal variations in their configuration pertaining to the count and depth of inference layers. The adjustment of these layers was performed manually to attain superior learning outcomes. 'DisplacementCNN' yields two final output images, each with dimensions "$h \times w$," which correspond to the predicted displacement components "$U$" and "$V$". Meanwhile, StrainCNN generates three output images, also of size "$h \times w$," each depicting one of the three plane strain components: "$\varepsilon_{xx}$", "$\varepsilon_{yy}$" and "$\varepsilon_{xy}$" The training dataset consists of images formatted to the size of *256 × 256*, as described in the following section.

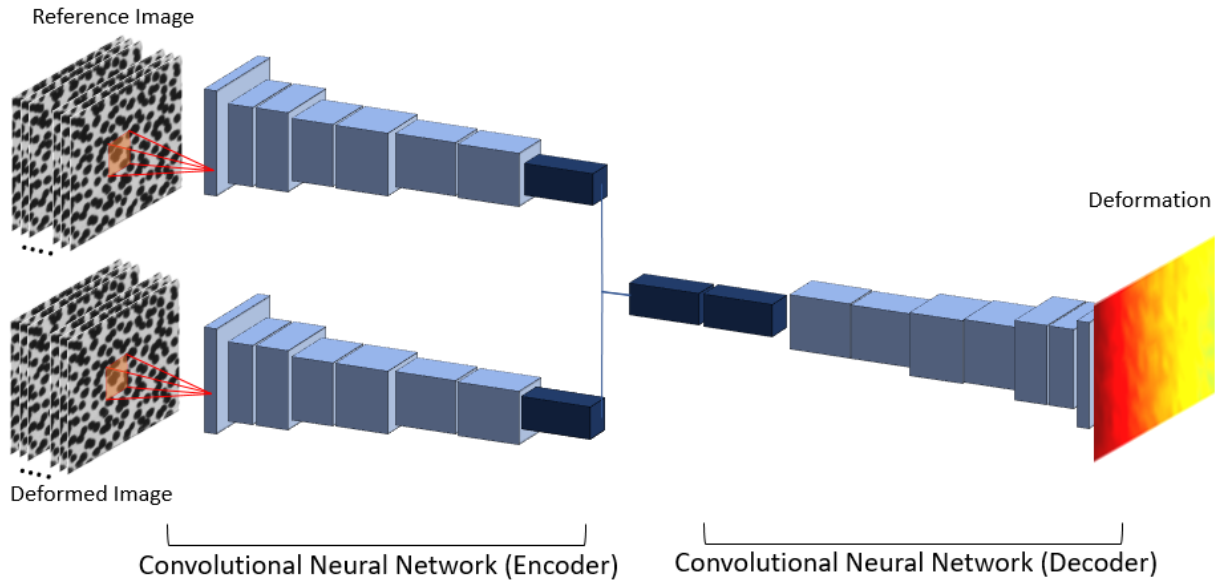

*Figure 2. The overall workflow of the proposed CNN for the deformation fields extraction from 2D images*

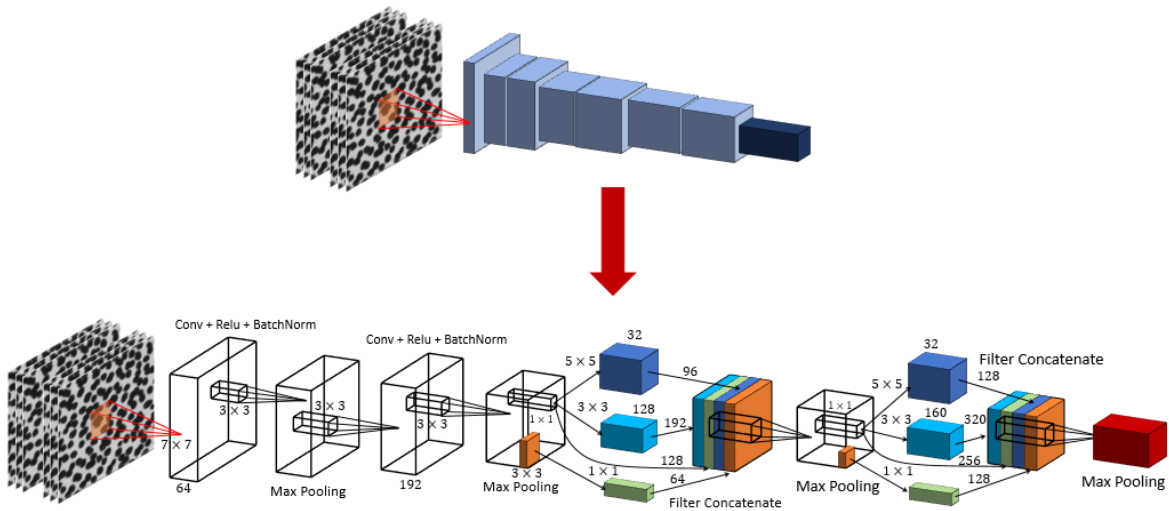

*Figure 3. A Comprehensive detail of the deep CNN to extract full-field deformation fields*

This research innovates by progressively updating the displacement and strain fields through incremental calculations. The ROI is determined based on the evolving coordinates of the four corner points within the cumulative displacement field. This approach allows for effective tracking of significant deformations. The procedure is iteratively applied to each frame in the sequence, with the ROI being updated for each new pair of image inputs. 'DisplacementCNN' and 'StrainCNN' operate independently to predict displacement and strain fields directly from the raw image inputs, respectively. The advanced CNN architecture adopted in this study features a convolutional autoencoder design. This includes multiple layers of CNNs arranged in an encoder-decoder configuration, establishing a foundational model for the analysis.

In the encoder segment of the proposed CNN, the architecture incorporates two inception blocks, alongside two convolutional layers. This configuration is further complemented by the inclusion of one max-pooling layer and an average-pooling layer, forming a comprehensive structure for the encoder component. The decoder network is structured in the reverse order of the encoder and utilizes up-sampling layers to increase field resolution. The convolutional layer applies a convolution operation to the input channels and passes the results to the next layer, allowing the CNN to extract distinctive features from the input images. Each convolutional layer is followed by a batch normalization and *ReLU* layer. The output of the model is the computed deformation fields, which are encoded in a 256 by 256 for simulated dataset and $150 \times 150$ matrix for experimental dataset and represented as a contoured image. Each pixel in the image represents the deformation fields corresponding to the deformed speckled patterns. The research incorporated a $1 \times 1$ convolutional layer, commonly referred to as the bottleneck layer, to minimize computational complexity and the quantity of parameters. Strategically placed prior to larger kernel convolutional filters, such as $3 \times 3$ and $5 \times 5$ layers, this bottleneck layer effectively reduces the parameter count involved in each pooling feature process. The introduction of a bottleneck layer serves a dual purpose: it not only decreases parameter numbers but also deepens the network by adding more nonlinearity. This is achieved through the application of *ReLU* activation following each layer. Furthermore, traditional fully connected layers were substituted with an average pooling layer, leading to a significant reduction in parameter count, as fully connected layers typically involve a high number of parameters. Figure 3, illustrates that the network in question is capable of learning more profound features with a reduced parameter count, especially when compared to networks like *AlexNet* or *ResNet*. Additionally, this architecture demonstrates enhanced speed relative to *VGG* [31].

Both the 'DisplacementCNN' and 'StrainCNN' models implement a tailored encoder-decoder architecture, a design frequently employed in tasks involving the segmentation of high-resolution output images [32]. In the encoder phase, a series of convolution operations with kernel sizes of 3 and 5 and a stride of 2 are utilized. This progressively diminishes the size of the feature map while augmenting its depth. Such a configuration enables the CNNs to derive rich features from the constrained data present in the input image pair. Conversely, the decoder segment employs deconvolution operations to reverse the encoding process. This gradually enlarges the feature map and reduces its depth at each level, thereby reconstructing the high-resolution displacement or strain field from the condensed feature maps. To mitigate the issue of gradient vanishing, batch normalization is applied prior to the activation function in each convolutional (or deconvolutional) layer, which notably expedites the training process.

*2.2. Transfer learning*

In general, Deep CNNs are designed and trained for specific applications using datasets that are tailored to the target problem. The challenge, therefore, is to determine whether a pre-trained CNN model, which has demonstrated good performance in estimating deformation fields for a specific dataset, can be utilized, either in full or in part, to solve the present problem of resolving sub-pixel deformations. The application of transfer learning and fine-tuning is proposed as a viable strategy to address this challenge. Transfer learning is a widely adopted strategy in deep learning where pre-trained models are leveraged to solve new, related tasks. By utilizing pre-trained model parameters, transfer learning enables faster convergence and reduced requirements for labeled training data. In this approach, CNNs are fine-tuned on a combination of the pre-existing training set and a new, task-specific training set, with the goal of facilitating their practical application. In scenarios where the goal is classification, transfer learning typically involves substituting the final fully connected layer of the model and training this new layer, while retaining the original weights of the convolutional layers. This approach is based on the premise that, if the current dataset closely resembles the one used for initial training, the outputs of the convolutional layers should remain pertinent to the target problem. However, given that our experimental data significantly differs from the datasets utilized in training the pre-existing CNNs, we elected to implement a comprehensive fine-tuning process. This process involves updating the weights across all layers, not just the final one. The weights from the pre-trained model serve as the initial points for this fine-tuning process, applied to our specific dataset. Figure 4 presents a schematic representation of the model's training process, which begins with *Dataset A* and subsequently transfers the acquired knowledge to *Dataset B*, highlighting the model's capability for extrapolation (Figure 4).

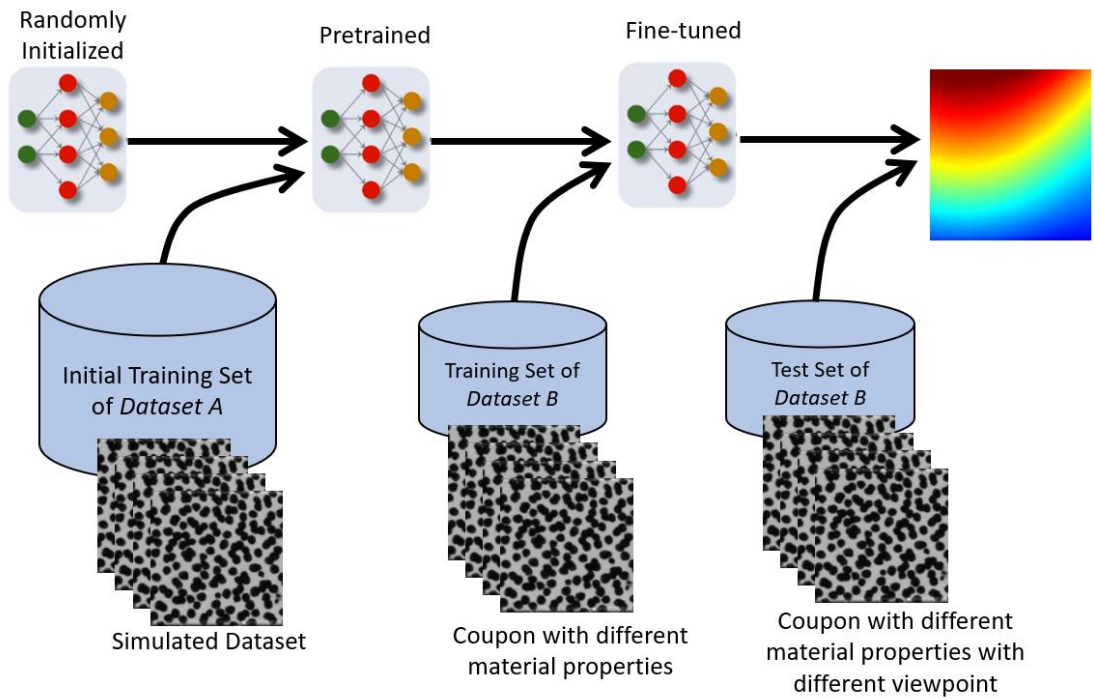

(a)

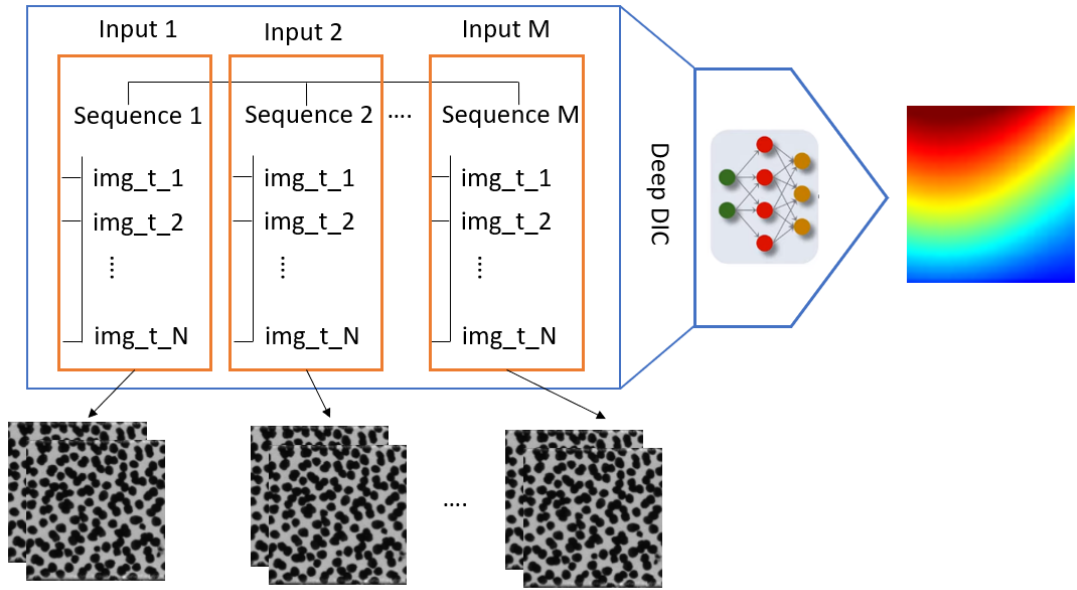

(b)

*Figure 4. (a) A diagrammatic process of training the model with Dataset A and then transferring the learned insights to Dataset B, (b) Inputs to the deep learning-based network*

*2.2.1. Efficacy Evaluation of transfer learning*

The efficacy of transfer learning is investigated from the speckle patterns. Subsequently, the network is trained using two distinct methods: the first method involves training the network from scratch with randomly initialized weights, while the second method leverages the transfer learning technique. Both methods employ the Adam optimization algorithm and utilize a learning rate of 0.0001. When training the network without utilizing transfer learning, the loss rate is often high and prone to large fluctuations. This instability suggests that the network is not in a steady state. However, when transfer learning is employed, the network only requires approximately 100 epochs to converge and the loss decreases in a more stable manner. Once the loss reaches 3%, fluctuations are minimal. These observations demonstrate that transfer learning can effectively accelerate the training process (Figure 5).

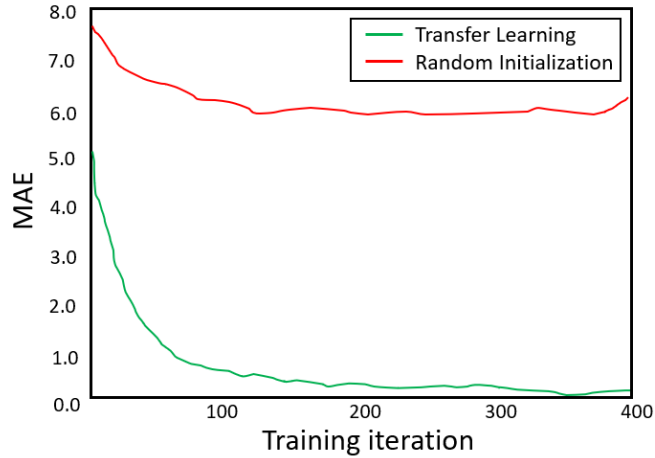

*Figure 5. Efficacy Evaluation of transfer learning, training epoch and loss.*

## 3. Data Preparation: Training/Testing Dataset

A deep learning algorithm typically requires a large amount of training data to achieve optimal performance. The reformulation of the structural deformation fields using a deep learning approach requires the creation of training datasets containing speckle images as well as the development of deep learning architectures able to process image sequences into corresponding deformation fields. Previous research in the context of optical flow estimation has utilized large datasets consisting of image pairs and ground truth optical flow, typically generated through the rendering of synthetic images. Despite the labor-intensive process of generating such datasets, several have been utilized in previous studies, however, these datasets are limited to large rigid or quasi-rigid motions and natural images [33]. Acquiring a significant number of speckle images experimentally can be excessively time-consuming. To overcome this challenge, simulated speckle pattern images are utilized with random deformations to meet the training requirements. Recent research has led to the creation of a dataset that consists of image pairs. These images simulate reference and deformed speckle patterns and are paired with their corresponding deformation fields [23, 34]. The dataset is representative of real-world images and deformations to ensure that the trained deep learning models can accurately infer the deformation field from novel image pairs not present in the training set

[23]. The proposed deep learning models was validated on two datasets, "*Dataset A*" from [23] comprising publicly accessible data, encompasses images of simulated speckled patterns with varying characteristics and "*Dataset B*" from [11] consists of experimental data. The batch normalization layer was used to standardize the inputs to each layer and the performance of the proposed deep learning models was verified through the simulated data and real experiments. The simulation results demonstrate its accuracy, while the real experiments show its robustness. The model training process utilize of *Dataset A* for model training, followed by the transfer of acquired knowledge to *Dataset B* to demonstrate the model's generalization and extrapolation capability.

It should be noted that the application of transfer learning can play a significant role in reducing overfitting during training. This is particularly pertinent given the limited size and scope of the currently available simulated speckle datasets (*Dataset A*) [23]. During the training phase with *Dataset B* [11], the weights of the model are updated, potentially at a lower learning rate. This allows the pre-trained model to serve as an initial weight framework for the new model. In the pre-trained encoder network, the last three fully connected layers were removed, preserving only the convolutional max-pooling layers. This modification was based on the understanding that these layers harbor more generalized features at the lower levels of the network, while encapsulating more complex, dataset-specific features at higher levels. Another rationale for the exclusion of the fully connected layers is their lack of spatial information encoding, which is crucial for identifying subtle motion patterns within the deep learning structure. Consequently, the model is directly linked to the final pooling layer of the deep learning architecture. This layer holds abundant spatial data, which the decoder can utilize to assess temporal aspects across consecutive frames. The methodology is divided into the following sequential steps:

1) *Data Acquisition for Model Training:* This step involves the collection of a foundational dataset, essential for the training of the deep learning model. It includes an array of simulated speckle images, characterized by diverse pattern parameters that are employed in the image rendering process. [23] (*Dataset A*).

2) *Data Gathering for Model Validation and Transfer Learning:* This phase entails assembling a dataset consisting of experimental data (referred to as Dataset B), which is then employed for implementing transfer learning. This dataset is pivotal for validating the model's performance and efficacy in diverse scenarios:

    - *Experimental data consists of a steel coupon specimen.*
    - *Experimental data consists of a steel coupon specimen with existing damage.*

3) *Identification of Full-Field Deformation on the Existing Dataset:* The deep learning methods proposed in this study will be utilized to determine the deformation fields of structures from the gathered images, constituting the previously mentioned Dataset A. This application aims to demonstrate the effectiveness of the deep learning approach in accurately capturing the deformation characteristics from the dataset.

4) *Selection of Deep Learning Architecture:* The chosen deep learning architecture incorporates transfer learning for the extraction of full-field deformation fields from constrained datasets. Through the implementation of transfer learning, the selected deep learning model is adept at applying insights gained from previously trained models. This enables it to extract deformation fields effectively and accurately, even when faced with limited data availability.

5) *Optimization through Hyperparameter Selection:* The optimization of the proposed deep learning model involves fine-tuning various hyperparameters. These include the number of convolution layers, kernel size, stride size, and the specific types of pooling layers used. The careful adjustment of these hyperparameters is crucial to ensure the model's optimal performance and efficiency in processing the data flow.

6) *Application of Transfer Learning to Novel Structures:* The concept of transfer learning is employed to broaden the scope and validate the effectiveness of the proposed deep learning methodologies on a complex structural system, as represented by Dataset B. This process of extension and validation is designed to yield deeper understanding and insights into the practical utility and the broader applicability of the deep learning approach under consideration.

*3.1. Simulated Speckle Images for Training Dataset (Dataset A)*

For the accurate estimation of minute subpixel deformations, the creation of a novel dataset, referred to as the Speckle dataset, is essential. The process of constructing this simulated speckle dataset involves several key stages: the generation of speckle reference frames, the establishment of displacement field parameters, and the creation of corresponding deformed frames. Each of these steps is elaborated upon in the reference, providing a comprehensive overview of the dataset construction methodology [23]. The dataset is composed of pairs of images that simulate reference and deformed speckle patterns, accompanied by their respective deformation fields. To guarantee the efficacy of the resulting CNN in deducing the deformation field from novel image pairs, not included in the training set, the dataset needs to closely resemble actual images and deformations. In practical experiments, the initial stage often involves prepping the specimen, typically through spray-painting to deposit black droplets on a white surface. This procedure is emulated in the dataset creation using the speckle generator method described in [35]. The speckle generator employed in this process overlays small black disks onto an image at random, creating synthetic speckle patterns that closely resemble real-life counterparts. Using this tool, reference frames with dimensions of $256 \times 256$ pixels were produced, incorporating a variety of settings or parameters as detailed in [35]. Further information about the characteristics and specifics of these simulated speckled images is available in [23].

*3.2. Experimental Design for Validation of Transfer Learning Dataset (Dataset B)*

*3.2.1. Specimens with different material, geometry, and Speckle patterns*

The DIC experimental setup was established to gather data for the fine-tuning and validation of the trained deep learning model. This laboratory-scale investigation involved testing coupon specimens under identical displacement-controlled tensile loading and boundary conditions. The captured images had a resolution of $1152 \times 648$ pixels and were segmented into 28 grids, each measuring $150 \times 150$ pixels, within the ROI. A subset image of $150 \times 150$ pixels, centered on the grid, was utilized for training purposes, leading to the generation of a total of 28000 processed data points. The specimens employed in this study were configured in five distinct forms, as illustrated in Figure 6 and detailed in the following descriptions:

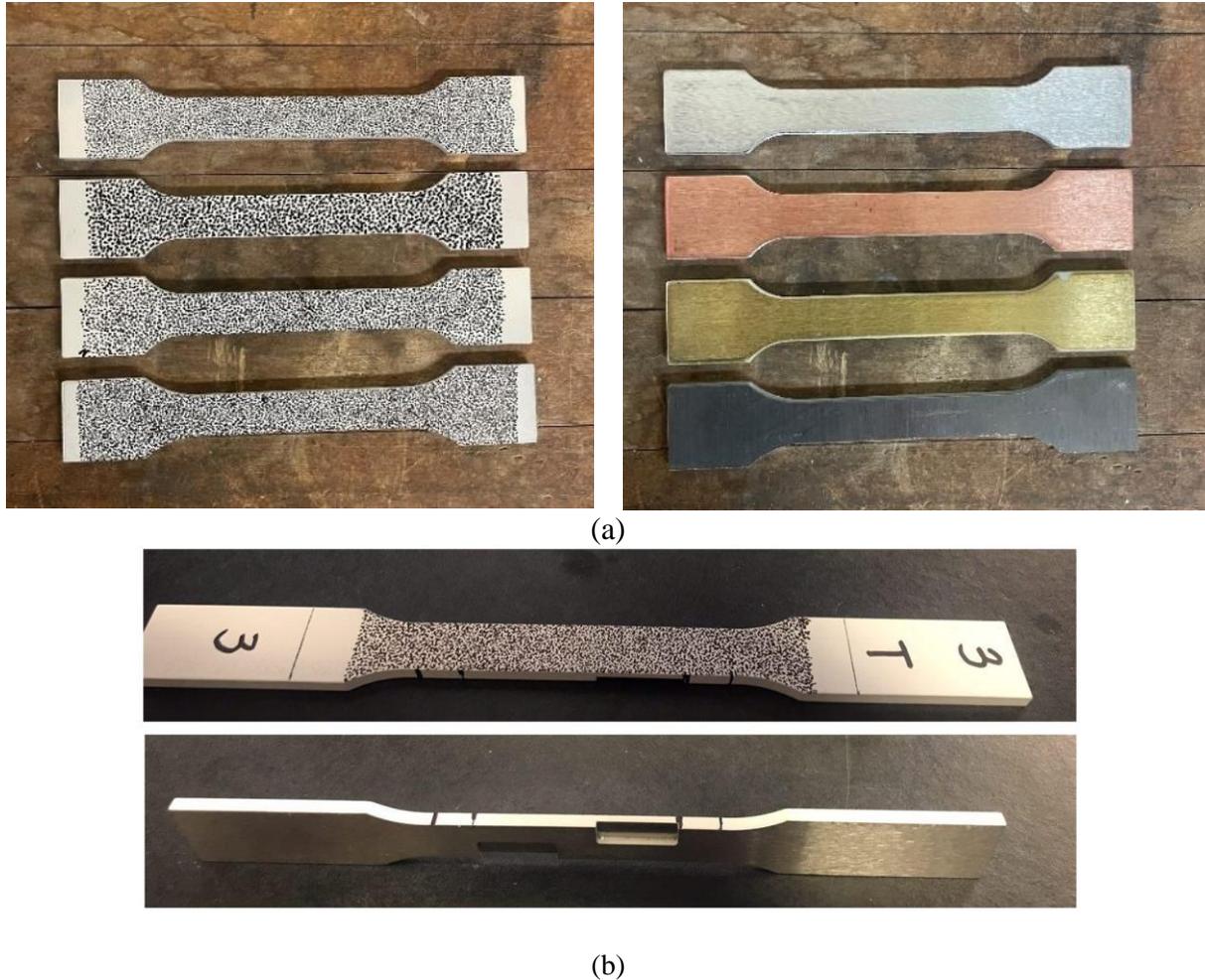

(a)

(b)

*Figure 6. (a) Coupon specimen components with different material properties consists of steel, copper, bronze and aluminum, (b) Coupon specimen with damaged region on the back side.*

### 3.2.2. Quasi-Static Mechanical Testing

In the context of Dataset B, this research involved conducting a sequence of quasi-static tensile tests on structural coupon specimens. These tests were integral to the fine-tuning and validation of the deep learning methodology proposed. Executed under uniaxial tensile loading, the tests remained within the elastic limits of the specimens. This testing arrangement produced a range of strain fields, both uniform and non-uniform, encompassing longitudinal, transverse, and shear strain components. Additionally, it yielded in-plane and out-of-plane displacement fields, including longitudinal and transverse components. The details of the experimental setup, along with the commercial DIC measurement system used, are illustrated in Figure 7. This figure also displays the ROI, within which the results obtained from the deep learning analysis were compared against measurements from the DIC system. The process of experimental validation entailed conducting simple tension tests on seven structural coupon specimens. These tests adhered to the methodology outlined in ASTM E8 standard [36], ensuring a standardized approach for the validation process. The DIC measurements in this study were conducted using two different camera setups. The first setup comprised a camera equipped with a 5-megapixel charge-coupled device (CCD) image sensor and a 12 mm optical lens. This camera was utilized for 2D DIC, enabling the extraction of in-plane deformation fields. The second setup involved a pair of stereo digital cameras, also equipped with 5-megapixel CCD image sensors, which were used for 3D-DIC. This allowed for capturing different perspective images of the specimens. The cameras were positioned two feet away from the coupon specimens. A stereo calibration of the system was conducted using a checkerboard pattern. The analysis and processing of the collected data were performed using Vic-2D and Vic-3D, which are commercially available DIC software products from Correlated Solutions Inc. [37]. Before commencing the tests, the surfaces of the specimens were coated with a layer of flat white paint and subsequently marked using a fine-tip permanent marker to create a dense, random speckle pattern, essential for the correlation process. In the post-processing phase, the speckle pattern within the ROI was segmented into smaller, rectangular windows or subsets. These subsets contained distinctive speckle patterns, which were ideal for template matching and tracking purposes. The patterns identified in each image were then correlated with those on a grid defined by a specific "*Step*" size, and their new positions in subsequent images were determined. The selection of an appropriate subset size and Step size is influenced by factors such as the speckle pattern, lens magnification, and imaging distance. This selection process is typically refined through direct experimentation during the post-processing stage. In this study, the subset size was chosen based on the dimensions of the speckle patterns, with a Step size of 7 pixels being utilized, as illustrated in Figure 7. Further information on the DIC setup can be found in the authors' previous works [9, 11-14, 38-41].

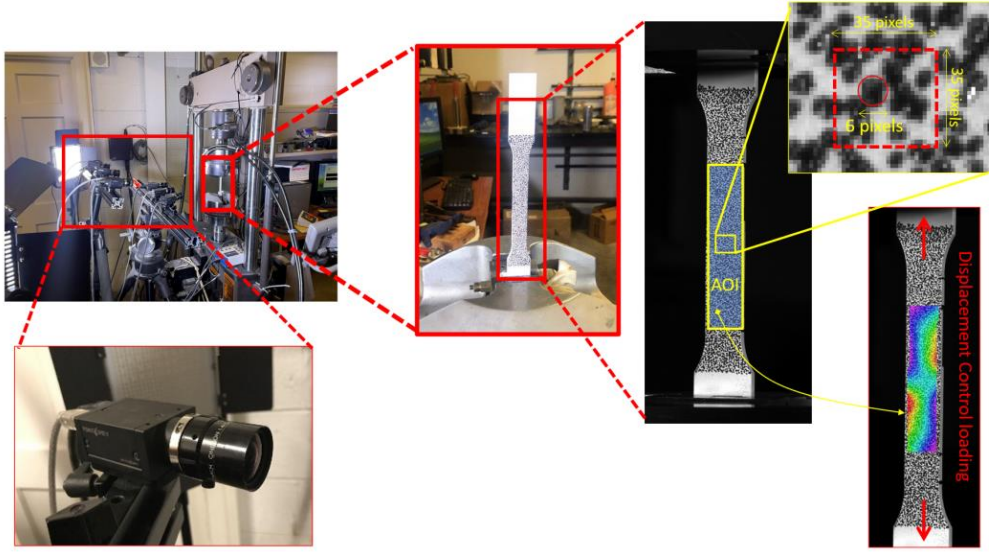

*Figure 7. Experimental DIC setup configuration (one set of stereo-paired cameras)*

## 4. Training Details, Network Implementation and Evaluation Metric

In this research, the training of the models was conducted using the Adam optimization algorithm, selected for its rapid convergence rate. This rate is enhanced by the algorithm's capacity to adaptively modify the learning rate in response to the current gradient. The hyperparameters $\beta_1 = 0.9 \ and \ \beta_2 = 0.999$, as recommended by Kingma and Ba in their 2015 study [42] were implemented. Additionally, a batch size of 16 and a relatively high weight decay of 0.1 were employed. The training regime also included a linear learning rate warmup followed by a decay phase. The dataset was partitioned into three segments: 75% for training, 10% for validation, and 15% for testing. To assess the performance of the proposed deep learning architectures, the Mean Absolute Error (MAE) metric [43] was utilized. This metric, which was also applied as the loss function during training, quantifies the absolute discrepancy between the predicted values and the actual ground truth, as delineated in equation (1).

$$MAE = \frac{\sum_{i=1}^{n}|\hat{y}_i - y_i|}{n} \qquad (1)$$

In the equation, $\hat{y}$ represents the predicted value, y denotes the actual ground truth value, and *n* signifies the sample size, which in this context is the number of sequences in the dataset. The development of the proposed CNN architectures was facilitated using TensorFlow, along with the high-level neural network API, Keras, to expedite the process. The models were initially trained on Dataset A, applying Xavier's method for initialization. For the minimization of the loss function, the Adam optimization technique was employed, configured with a momentum value of 0.9. All weight parameters in the model were subject to L2 regularization. Additionally, a dropout ratio of 0.5 and a regularization weight of 0.0005 were implemented. The effectiveness of the model was gauged using the MAE metric, comparing the predicted deformation fields against their actual counterparts. The choice of the Adam optimizer was based on its demonstrated advantages over alternative optimization methods. The learning rate was set at 0.0001 across a total of 400 training epochs. To counteract overfitting, the model's performance was continually assessed using a validation dataset. The training of the network was conducted on a commercial laptop equipped with 16GB of RAM and a 4GB video RAM GTX 860 M GPU. The software environment included Keras version 2.2.4 and TensorFlow GPU version 1.13.1, operating on a backend supported by CUDA 10.0 toolkit and cuDNN 7.5.

Figure 8 presents the MAE values of the developed model, illustrating its exceptional performance. For the 'DisplacementCNN' model, the learning rate commenced at 0.0001 and was decreased by a factor of 100 post 220 epochs. After 200 epochs of training, the error on the validation set for 'DisplacementCNN' was observed to be less than 0.01. In the case of the 'StrainCNN' model, the learning rate started at 0.001 and was lowered to 1e-5 following 220 epochs. The training for 'StrainCNN' concluded at epoch 400, with a validation error noted at 0.06. The progress of convergence for both 'DisplacementCNN' and 'StrainCNN' is depicted in Figure 8, and their respective MAE predictions are consolidated in Table 1. The term 'generalizability' pertains to the ability of the model to generate accurate outcomes on data it has not encountered before, but which falls within its training input domain. The network iteration exhibiting the minimal validation loss was preserved and subsequently fine-tuned on various datasets for predictive and testing applications. The detailed layer configuration of the proposed network is outlined in Table 2.

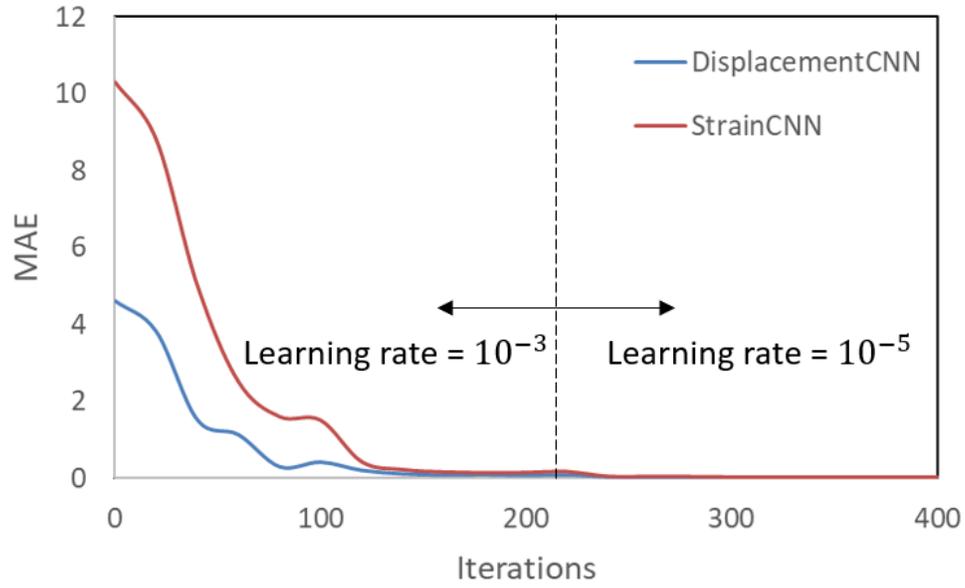

*Figure 8. Convergence history of the deep learning models*

*Table 1. Performance of the deep learning models on the validation and test sets*

| Prediction error | DisplacementCNN (pixel) | StrainCNN (%) |
|---|---|---|
| Validation set | 0.03 | 0.06 |
| Test set | 0.07 | 0.08 |

*Table 2. The layers configuration of the proposed network for 3D*

| Layer | Configuration | Name | Modules |
|---|---|---|---|
| 1 | 64 kernels - kernel size (7×7) | CNN 1 | |
| 2 | | Batch Normalization | |
| 3 | Pooling size (3×3) - Stride (2×2) | Max Pooling | |
| 4 | 192 kernels - kernel size (3×3) | CNN 2 | |
| 5 | | Batch Normalization | CNNs |
| 6 | Pooling size (3×3) - Stride (2×2) | Max Pooling | |
| 7 | Filters - 128,128,192, 32, 96, 64 | Inception Layer 1 | |
| 8 | Pooling size (3×3) - Stride (2×2) | Max Pooling | |
| 9 | Filters - 256, 160, 320, 32, 128, 128 | Inception Layer 2 | |
| 10 | Pooling size (3×3) - Stride (2×2) | Max Pooling | |

## 5. Performance of the deep learning models on a tensile test sample of steel

Upon completion of tests on *Dataset A*, as detailed in [44], the research focused on assessing the trained network's extrapolation capabilities using a novel dataset, referred to as *Dataset B*. This dataset was derived from video recordings of coupon specimens subjected to loading, as previously outlined. The network's extrapolative potential was gauged based on its predictive accuracy on *Dataset B*, which lay outside its initial training scope. A groundbreaking approach was adopted to showcase the capacity of the proposed deep learning models to extend their applicability to this new dataset. The implementation of this approach involved the following procedural steps:

- Initially, the model under consideration was trained utilizing *Dataset A*, as referenced in [23]. This dataset comprised images of simulated data, serving as inputs for the network to derive the deformation fields as outputs. Post-training, the model demonstrated proficiency in accurately predicting the deformation fields of structures represented by simulated images, encompassing a range of varying parameters.

- Subsequently, transfer learning was employed to refine the model's weights for *Dataset B*. This dataset consisted of images derived from experimental data, used for predicting deformation fields, as previously delineated. Following this fine-tuning process, the model exhibited the capability to accurately predict the deformation fields of structures, specifically those comprising simple coupon specimens.

- For *Dataset B*, transfer learning was implemented across various scenarios involving video frames captured from different viewpoints of the coupon specimen. The model was then tested on alternate viewpoint video frames of the same specimen. These distinct scenarios, summarized in Table 3 were utilized to fine-tune and validate the pretrained deep learning models, ensuring their adaptability and accuracy across varying perspectives.

- To authenticate the model's performance on a completely independent dataset, artificial damage was induced in the rear region of the specimens to replicate a damaged state, thus generating a distinct structural variant. This "damaged" dataset, not utilized in any training or fine-tuning phase of the model, served as a novel testbed. The refined model was subsequently assessed using this new dataset, which encompassed varied perspectives of the damaged coupon component. The specifics of these different views are outlined in Table 3.

*Table 3. Different scenarios to fine-tune and validate the proposed deep learning models*

| Scenarios | Same material properties |
|---|---|
| 1 | Fine-tune the pretrained model using 2D speckle images consists of steel coupon and test using 3D speckle images consists of steel coupon |
| 2 | Fine-tune the pretrained model using 2D speckle images consists of steel coupon and test using 3D speckle images consists of bronze coupon |
| 3 | Fine-tune the pretrained model using 2D speckle images consists of steel coupon and test using 3D speckle images consists of aluminum coupon |
| 4 | Fine-tune the pretrained model using 2D speckle images consists of steel coupon and test using 3D speckle images consists of copper coupon |
| 5 | Fine-tune the pretrained model using 2D speckle images consists of steel coupon and test using 3D speckle images consists of steel coupon with damage region |

*5.1. Scenarios 1 – Configuration with 2D Speckle patterns dataset training/validation*

In this research, a comprehensive collection of 7200 images from tensile tests on coupon specimens, each featuring different material properties, was compiled. These images were crucial for comparing and assessing the performance of the proposed deep learning models against the results obtained from DIC VIC-3D. In the initial scenario, the predicted displacement and strain fields at four distinct time frames were analyzed. Their comparative performance is depicted in Figure 9 and Figure 10. The first frame, captured at time $t_1$, illustrated the early stages of deformation. The subsequent frames at $t_2$ and $t_3$ displayed localized deformation post-yielding. Notably, the VIC-DIC results at $t_2$ indicated the beginning of tearing in the top middle region of the ROI, attributed to the elongation of speckle patterns, resulting in dots exceeding the predefined step size. The final frame, taken at $t_4$, was near the test's conclusion and showed the onset of cracks, as evidenced by the tearing of speckle patterns on the sample.

This study's findings offer crucial insights into the effectiveness and accuracy of deep learning-based DIC, especially when compared to VIC-3D, in predicting displacement and strain fields at various deformation stages during tensile tests. The goal was to capture local deformations with high precision and stability for both displacement and strain fields. To this end, the output resolution for both fields was maintained equal to the input image size. The subset and step sizes recommended by the software, set at 35 and 7 respectively, were adopted. The output size produced by VIC-3D was approximately one-seventh of the original image size, leading to a uniform output size of $64 \times 64$ for all predictions. For a direct comparison, the results from VIC-3D were interpolated to align with the dimensions of the deep learning-based DIC predictions.

In Figure 9, a comparison of displacement predictions in the longitudinal direction is presented. The results demonstrate a high level of consistency and similarity in the displacement field predictions generated by 'DisplacementCNN' compared to VIC-3D. Specifically, the absolute magnitude and spatial distribution of the displacement field predictions are found to be in good agreement. However, for time instance $t_4$ the VIC-3D results display some quality fluctuations, as indicated by a large blank spot in the speckle image. In contrast, DisplacementCNN is not impacted by this pattern variation due to its training set's inclusion of different quality speckle images.

Figure 10 presents a comparative analysis focusing on the predicted strain $\varepsilon_{yy}$. Similar to the previous observations, there is a substantial correlation between the predictions from the two methods, with comparable magnitudes and spatial distributions. However, notable discrepancies are evident in the VIC-3D results at time $t_4$, particularly in areas of significant deformation where the speckle patterns start to tear, leading to invalid predictions. In these regions, 'StrainCNN' consistently yields plausible outcomes, underscoring a key advantage over traditional DIC techniques. 'StrainCNN' exhibits enhanced robustness in dealing with various pattern changes, including those around edges and in torn speckle areas. This attribute is especially crucial for materials undergoing extremely large strains. Moreover, 'StrainCNN' demonstrates superior spatial resolution in detecting localized strain concentrations at the sample's center in frame $t_4$. This ability to accurately capture such nuances aligns more closely with optical observations, as depicted in the figure. This precision in detailing strain concentrations exemplifies StrainCNN's advanced capability in strain field analysis under complex deformation scenarios.

Additionally, to demonstrate the proficiency of deep learning-based DIC in attaining superior resolution and computational efficiency, comparisons and discussions were made with VIC-3D. In traditional DIC methods, the chosen subset and step sizes significantly influence both the output resolution and the computational time. It is critical that the subset size is sufficiently large to encompass enough pattern features for effective correlation, although this consideration has a direct impact on spatial resolution. The step size is a direct determinant of both the output size and the computation time; notably, reducing the step size by half typically results in a fourfold increase in calculation time.

The DIC method requires the setting of various parameters to estimate displacement, presenting a significant challenge in selecting an optimal subset size. This size must be carefully chosen to accurately estimate the displacement of a specimen, considering its pattern or deformation. An inadequately selected subset size can result in erroneous displacement estimates, especially in cases of large displacements. Additionally, the strain measurement of a specimen is influenced by factors like the elastic and plastic regions and the direction of loading, further underscoring the criticality of optimal subset size selection in the DIC technique. To address the limitations inherent in traditional DIC methods, the proposed innovative deep learning based DIC approach employs deep learning algorithms to automatically determine the most suitable subset size for displacement estimation, thereby facilitating precise strain measurement. In this deep learning DIC framework, both 'DisplacementCNN' and 'StrainCNN' conduct pixel-level predictions, ensuring the output image size is always identical to the input size. The computation time is proportional to the input image size but remains unaffected by output resolution or the quality of the speckle pattern. Additionally, once fully trained, the model requires fewer adjustments to settings. For instance, in a tensile test with 1800 frames, using VIC-3D with a subset size of 35 and a step size of 7, the computation time is approximately 624 seconds. In contrast, the trained deep learning DIC model completes the calculation of both displacement and strain fields in just 1.2 seconds, averaging 0.67 milliseconds per frame. This time includes both image file loading and the computation process, illustrating a significant efficiency advantage.

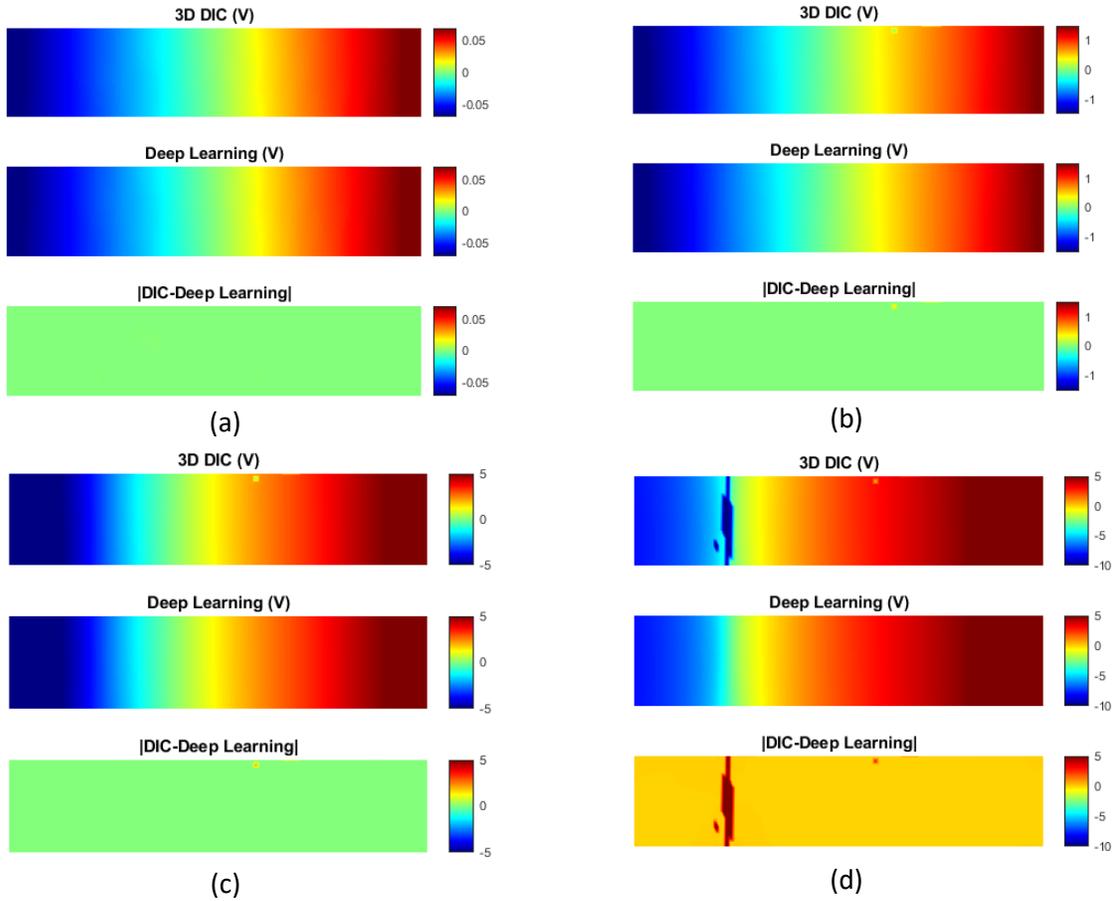

Figure 9. Comparison of displacement prediction from DisplacementCNN and VIC-3D for tensile testing on a steel sample for scenario 1, a) $t_1 = 100$ (b) $t_1 = 500$ (c) $t_1 = 1500$ (d) $t_1 = 2000$

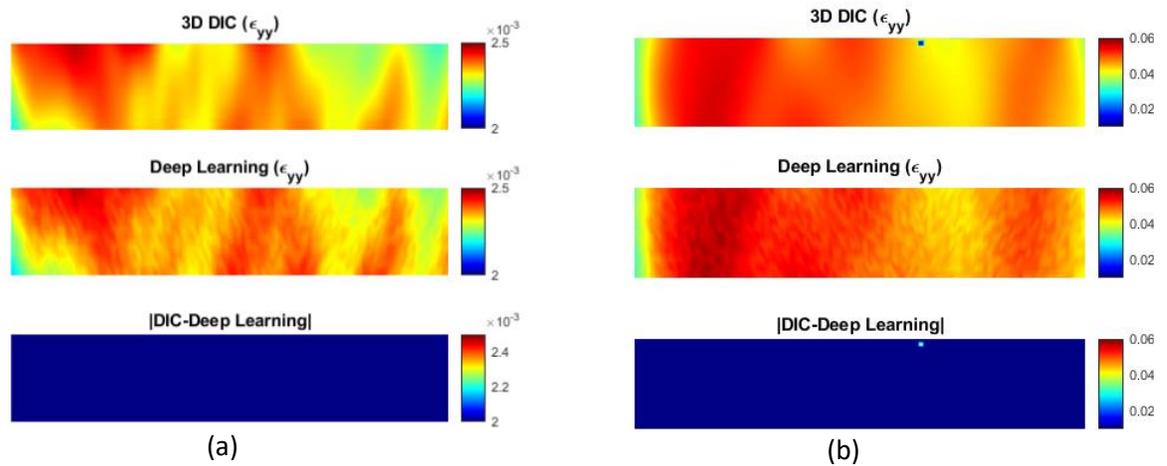

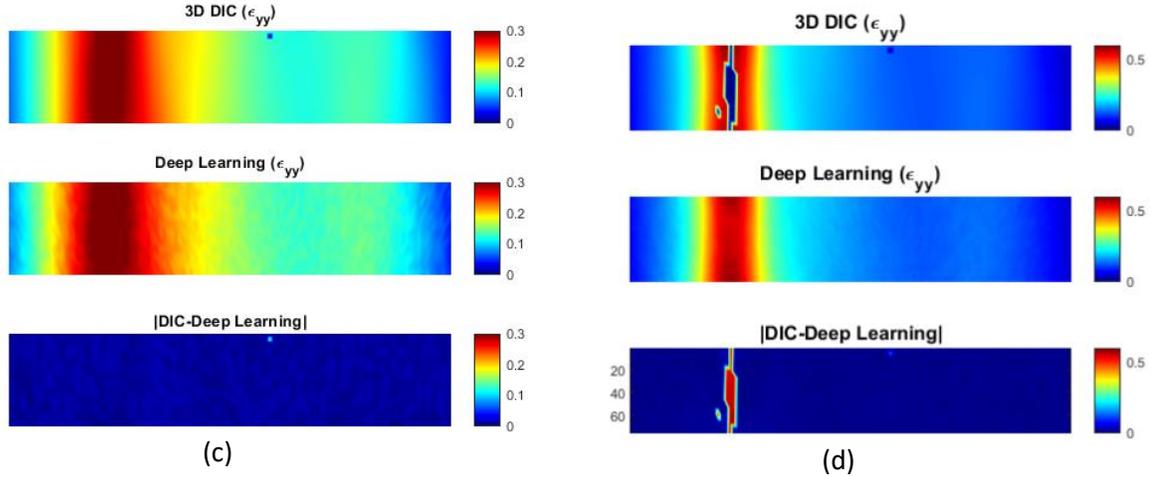

*Figure 10. Comparison of strain prediction from StrainCNN and VIC-3D for tensile testing on a steel sample for scenario 1, (a) $t_1 = 100$ (b) $t_1 = 500$ (c) $t_1 = 1500$ (d) $t_1 = 2000$*

Scenarios 2 to 4 in this study focus on evaluating the transferability and generalizability of the proposed deep learning network in predicting deformation fields of coupon specimens with varied properties. This analysis assumes that the model, initially fine-tuned with 2D images of steel specimens, is tested on 3D images of specimens possessing different characteristics. The outcomes, as depicted in Figure 11, Figure 12 and Figure 13 reveal that the deep learning network adeptly utilizes the most pertinent spatial features, effectively retains spatial information of visual features, and focuses on the critical temporal features embedded in key frames. These results attest to the model's ability to be transferable, generalizable, and extrapolatable. It demonstrates the network's proficiency in accurately extracting deformation field properties from new and previously unseen videos, which are fine-tuned using 2D images and tested with 3D images. This capability highlights the model's adaptability and its potential for broad application in various structural analysis scenarios.

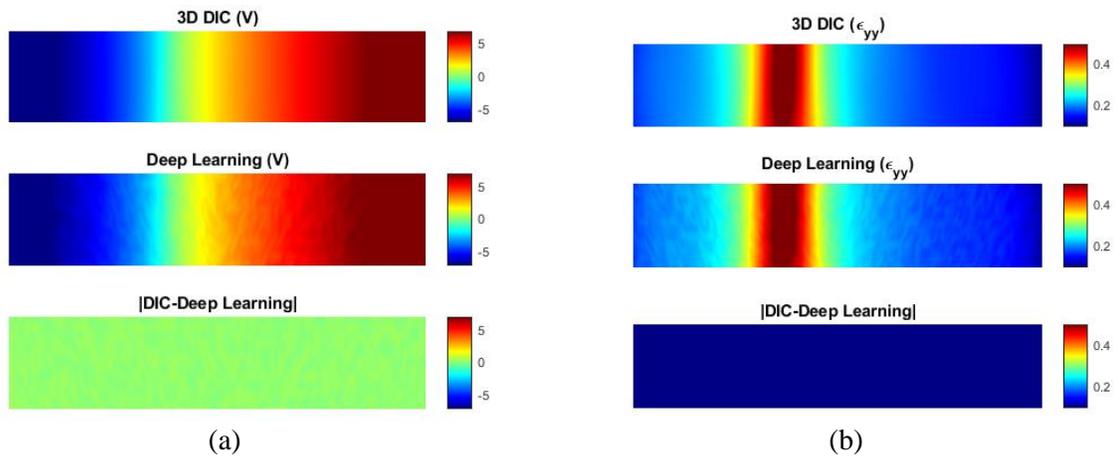

(a)            (b)

*Figure 11. Comparison of displacement and strain prediction from DisplacementCNN and Strain CNN with VIC-3D for tensile testing on a bronze sample*

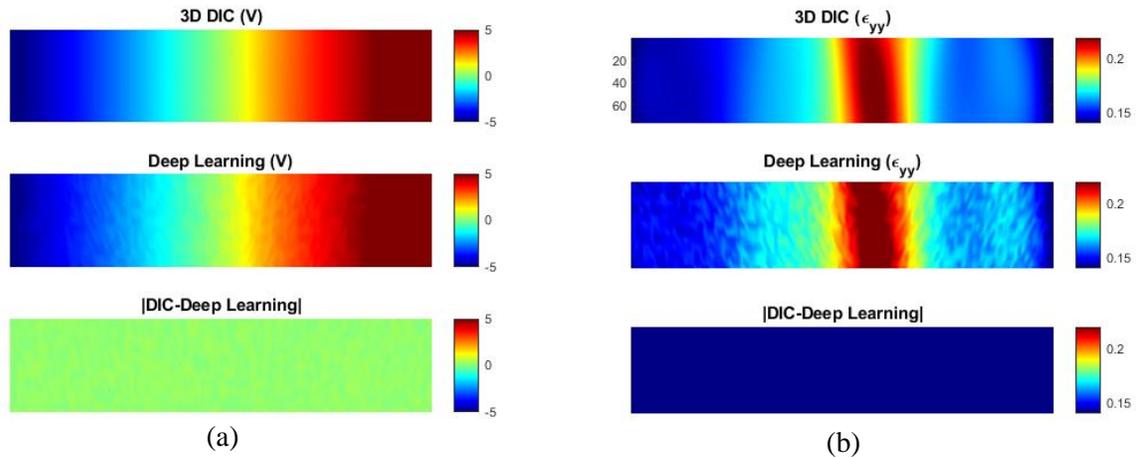

(a)            (b)

*Figure 12. Comparison of displacement and strain prediction from DisplacementCNN and StrainCNN with VIC-3D for tensile testing on an Aluminum sample.*

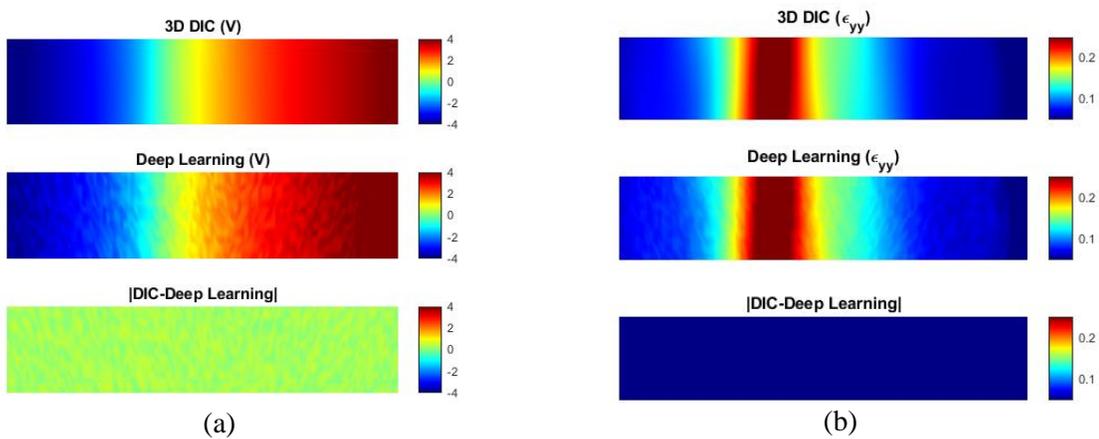

(a)            (b)

*Figure 13. Comparison of displacement and strain prediction from DisplacementCNN and StrainCNN with VIC-3D for tensile testing on a Copper sample.*

*5.2. Scenarios 5 – Fine-tune with 2D images and test with 3D images consist of damage*

Figure 14 further demonstrates the capability of the transferred model to identify the deformation field of a different, yet analogous structure, such as a damaged coupon specimen. The calculated errors for the deformation field estimations substantiate the model's proficiency in predicting deformation fields of previously unobserved damaged structures, thereby underscoring its extrapolative potential. These outcomes, in conjunction with earlier observations, confirm that the proposed deep learning network possesses the qualities of transferability, generalizability, and extrapolation. This effectiveness is rooted in the network's adeptness at harnessing the most pertinent spatial features, maintaining the integrity of spatial information within visual features, and focusing on essential temporal features concealed within keyframes. This combination of capabilities enables the model to accurately extract deformation field properties from new and unseen video data.

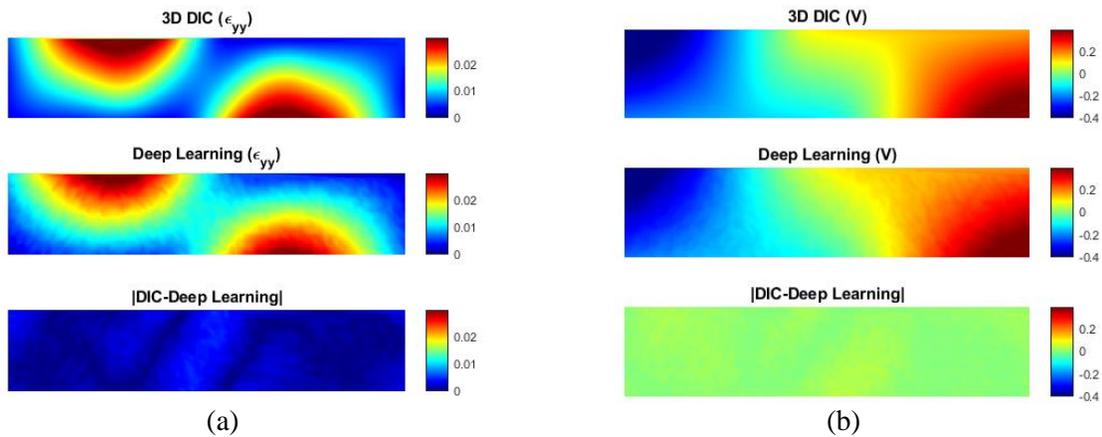

(a)          (b)

*Figure 14. Comparison of displacement and strain prediction from DisplacementCNN and StrainCNN with VIC-3D for tensile testing on a damaged steel sample.*

**Conclusion**

This research introduces an innovative deep learning model adept at capturing spatial-temporal information from image sequences. Utilizing the strengths of CNN, the model excels in extracting the most significant visual-spatial and temporal features from images. Experimental findings indicate that the proposed approach surpasses existing neural network-based estimators in predicting deformation fields from image sequences within current datasets. This trained network holds promise for real-time application in structural monitoring under load. Future research will concentrate on validating this technique on in-situ structures and expanding the training datasets to enhance the model's inferencing abilities. Once trained on an existing dataset, the deep learning model becomes a fully autonomous system capable of processing videos and deducing deformation fields. The presented deep learning algorithm for deformation field estimation embodies a vision-based method, demonstrating the utility of deep learning in deriving deformations directly from raw images. The algorithm leverages an existing dataset and applies transfer learning to extrapolate deformation fields beyond its training scope. It fine-tunes a pre-trained model to recognize new structural deformation fields with limited data. The potential of this algorithm extends to identifying other structural characteristics, such as mode shapes, provided these are incorporated into the training data, paving the way for future research with significant potential. Future efforts will focus on minimizing the error percentage by refining the deep neural network architecture and acquiring a more extensive dataset, thereby augmenting the model's precision and applicability.

**Acknowledgement**

We gratefully acknowledge the support of the National Science Foundation (NSF) under Award Number 2136724 for the project "EAGER: Adaptive Digital Twinning: An Immersive Visualization Framework for Structural Cyber-Physical Systems." We extend our sincere thanks to the NSF Division of Information and Intelligent Systems for their contributions to the advancement of this research. Special thanks are due to the Program Manager, Jumoke Ladeji-Osias, for her guidance and support throughout the project's duration, starting from October 1, 2021, to the estimated end on September 30, 2024. Additionally, we would like to thank the principal investigators—Devin Harris, Bradford Campbell, Panagiotis Apostolellis, and Jennifer Chiu—for their leadership and collaborative efforts. The total award amount of $300,000 has been instrumental in enabling the research conducted under this grant.